\documentclass[english]{cccconf}
\usepackage{amsmath,amsfonts}
\usepackage{algorithmic}
\usepackage{algorithm}
\usepackage{array}
\usepackage[caption=false,font=normalsize,labelfont=sf,textfont=sf]{subfig}
\usepackage{textcomp}
\usepackage{stfloats}
\usepackage{url}
\usepackage{verbatim}
\usepackage{graphicx}
\usepackage{cite}
\usepackage{multirow}
\usepackage{booktabs}
\usepackage{xcolor}
\usepackage{tabularx}
\usepackage{booktabs}
\usepackage{amssymb}
\usepackage[hidelinks, linkcolor=blue, anchorcolor=blue, citecolor=blue]{hyperref} 
\begin{document}

\title{LEO-RobotAgent: A General-purpose Robotic Agent for Language-driven Embodied Operator}

\author{Lihuang Chen\aref{zju}, Xiangyu Luo\aref{zju}, Jun Meng†\aref{zju}}

\affiliation[zju]{Zhejiang University, Hangzhou 310000, P.~R.~China
        \email{lihuangchen@zju.edu.cn; 22360566@zju.edu.cn; junmeng@zju.edu.cn}}


\maketitle

\begin{abstract}
We propose LEO-RobotAgent, a general-purpose language-driven intelligent agent framework for robots. Under this framework, LLMs can operate different types of robots to complete unpredictable complex tasks across various scenarios. This framework features strong generalization, robustness, and efficiency. The application-level system built around it can fully enhance bidirectional human-robot intent understanding and lower the threshold for human-robot interaction. Regarding robot task planning, the vast majority of existing studies focus on the application of large models in single-task scenarios and for single robot types. These algorithms often have complex structures and lack generalizability. Thus, the proposed LEO-RobotAgent framework is designed with a streamlined structure as much as possible, enabling large models to independently think, plan, and act within this clear framework. We provide a modular and easily registrable toolset, allowing large models to flexibly call various tools to meet different requirements. Meanwhile, the framework incorporates a human-robot interaction mechanism, enabling the algorithm to collaborate with humans like a partner. Experiments have verified that this framework can be easily adapted to mainstream robot platforms including unmanned aerial vehicles (UAVs), robotic arms, and wheeled robot, and efficiently execute a variety of carefully designed tasks with different complexity levels. Our code is available at \url{https://github.com/LegendLeoChen/LEO-RobotAgent}.
\end{abstract}

\keywords{Autonomous task planning, LLMs-based agent, embodied intelligence, general-purpose robotic intelligence, human-robot collaboration}

\footnotetext{
† Corresponding author: Jun Meng.}
\footnotetext{This work was supported by the ``Pioneer" and ``Leading Goose" R\&D Program of Zhejiang Province (No. 2024C01170), the National Natural Science Foundation of China (No. 52475033) and the Robotics Institute of Zhejiang University under Grant K11805 and K12404.
}

\section{Introduction}
Large language models (LLMs) and robotics are two highly prominent fields in current technology. In recent years, a growing number of outstanding works have begun to focus on the integration of these two domains. Endowed with powerful natural language understanding and logical reasoning capabilities, LLMs hold considerable research value in the field of robot task planning. The traditional robotics industry faces significant challenges in the field of task planning. For individual tasks, developers are required to manually design dedicated programs with specific logic, which suffer from poor flexibility and lack generalization capability. As task complexity and granularity increase, the program architecture inevitably becomes overly sophisticated and bloated. Furthermore, programs developed for specific tasks cannot be generalized or transplanted across different tasks. The emergence of LLMs has made us realize that this entity with strong reasoning capabilities can effectively replace the diverse and complex logics that we previously embedded in programs manually. Indeed, works over the past two years have incorporated LLMs into robot planning to varying degrees. References \cite{uavs} and \cite{robots} present and summarize numerous studies that have verified on different robot platforms, such as unmanned aerial vehicles (UAVs) and robotic arms, that LLMs can effectively enhance the planning performance of specific tasks, make program design more intuitive, and allow direct human-robot interaction.

Meanwhile, intelligent agent architectures represented by the ReAct framework\cite{react} have emerged and achieved remarkable development in the field of LLMs, aiming to address LLM task planning and execution problems in pure software applications. This is an exciting emerging subfield, indicating that LLMs also have the potential to be applied in robotics. This would allow LLMs, as the brain, to truly possess the body and tools to control robots performing various physical tasks. Based on the aforementioned findings, we argue that LLMs can provide a general foundational solution for high-level task planning of robots. Under this solution, the task planning system can break free from the constraints of manually predefined fixed rules and instead leverage the world knowledge and cognitive capabilities of LLMs for reasoning. It can autonomously adhere to task rules and reject the majority of meaningless trial-and-error attempts, thus evolving behavioral planning for diverse objectives in a self-driven manner. Ultimately, this enables robots to possess human-like divergent thinking and execution capabilities to address various tasks in different scenarios.

In this paper, we conduct in-depth research on the application of intelligent agent frameworks in the robotics domain. We propose a  streamlined and efficient general-purpose robotic agent framework, enabling LLMs to independently perform behaviors such as planning, action execution, and reflective adjustment based on user requirements and task information, ultimately achieving self-iterative control of robots to complete tasks. Human users can optionally intervene during task execution to implement interactive behaviors of varying degrees. Furthermore, through experiments, we present the usage suggestions of different technologies for task deployment and human-robot communication from the perspective of ordinary users in this framework. Inspired by existing excellent works, we test and compare the strengths and weaknesses of agent frameworks with different core principles and structures in our designed task scenarios, hoping to provide valuable references for researchers. In summary, our main contributions are as follows:

\begin{enumerate}
\item{We developed an elegantly structured general-purpose robotic intelligent agent framework called LEO-RobotAgent. The framework is equipped with human-robot interaction capabilities. We applied several prompt engineering techniques from the LLMs domain to explore the boundary of the task planning capabilities of LLM within this framework and identified the current problems and challenges faced by LLMs in complex task planning.}
\item{We provide a complete program system built around this framework. The system features a simple and easily configurable environment with visual and monitorable operation processes. Users can register various tools in the framework’s toolset module to meet customized requirements. With high flexibility, the framework effectively lowers the threshold for human-robot interaction.}
\item{We verified that this framework can be applied to different types of robots to accomplish tasks of varying complexity across diverse scenarios, exhibiting general applicability and high generalization.}
\end{enumerate}

\section{Related Work}
\textit{1) Large Models for Auxiliary Tasks:} Large Models have been applied to varying degrees across diverse single scenarios or robot types. They can integrate semantic information from multiple sources, such as sensors, visual models, and task objectives \cite{assist1}, and provide critical information like positional relationships for visual image understanding to assist robotic arms in task execution \cite{assist2}. Additionally, LLMs can be incorporated into the control loop of Unmanned Aerial Vehicles (UAVs) to generate dynamic and safe flight recommendations at low frequencies \cite{assist3}, or serve as high-level planners to develop action plans for disaster relief scenarios \cite{assist4}. Vision-Language Models (VLMs) can also act as downstream models of visual systems to further analyze urban patrol scenarios \cite{assist5}.

\textit{2) LLMs in Task Planning and Execution:} Representative works \cite{codegeneration1}, \cite{codegeneration2}, \cite{codegeneration3} have highlighted the significance of program generation for task execution, where executable code translates LLMs’ action plans into direct execution to accomplish short-duration, simple tasks. \cite{vla1}, \cite{vla2} adopt a straightforward approach by enabling LLMs to output robotic arm actions for completing basic task steps. In contrast, \cite{agent1}, \cite{agent2}, \cite{agent3}, \cite{agent4} truly empower LLMs to act as agents for task planning and step-by-step execution: \cite{agent2}, \cite{agent3} employ multiple LLMs with divided responsibilities (e.g., planning and supervision), while \cite{agent4} designs a complex upstream-downstream framework that allows LLMs to perform multi-level planning and establish effective feedback channels for error correction.

\textit{3) Other Inspiring Works:} \cite{other1} trains a multimodal large language model for embodied reasoning tasks in robotics. \cite{other2} provides comprehensive task benchmarks for Vision-Language-Action models(VLAs). Unlike these end-to-end VLAs that directly map sensor inputs to actions and require massive domain-specific robotic data for training, our tool-based framework leverages the zero-shot reasoning capabilities of off-the-shelf LLMs. This significantly lowers the barrier to deployment and enhances interpretability, though its performance depends on the capabilities of the toolset. Furthermore, the Qwen series models proposed in \cite{qwen2.5} and \cite{qwen3} offer excellent model options for our designed framework. \cite{other3} develops a robust system for hardware-software co-design of LLMs on UAVs. \cite{other4} identifies remaining limitations of LLMs in task planning, understanding, and reasoning through benchmark testing and analysis.

Current research endeavors face several limitations and challenges. Most works focus solely on a single robot type to execute a single task category. In many studies, LLMs only participate in task planning to a limited extent, resulting in poor algorithmic autonomy. Some program-generation-based frameworks often require rigorous validation, and the generated programs themselves cannot perform tasks that demand natural language understanding. Additionally, complex LLM-centric frameworks constructed in numerous works typically introduce issues such as difficult debugging and low stability. Furthermore, most algorithms do not support secondary human-robot interaction after task deployment.

\section{Methodology}

\begin{figure}[!t]
\centering
\includegraphics[width=3.0in]{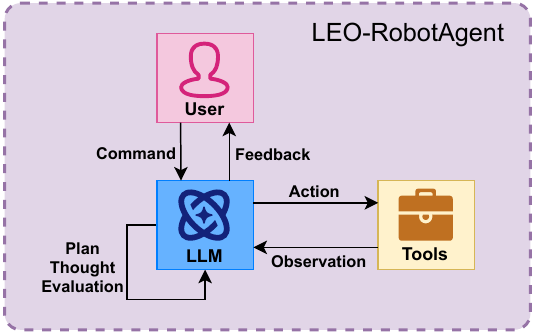}
\caption{Basic schematic of LEO-RobotAgent.}
\label{fig_1}
\vspace{-0.5cm}
\end{figure}

\begin{figure*}[!t]
\centering
\includegraphics[width=6.2in]{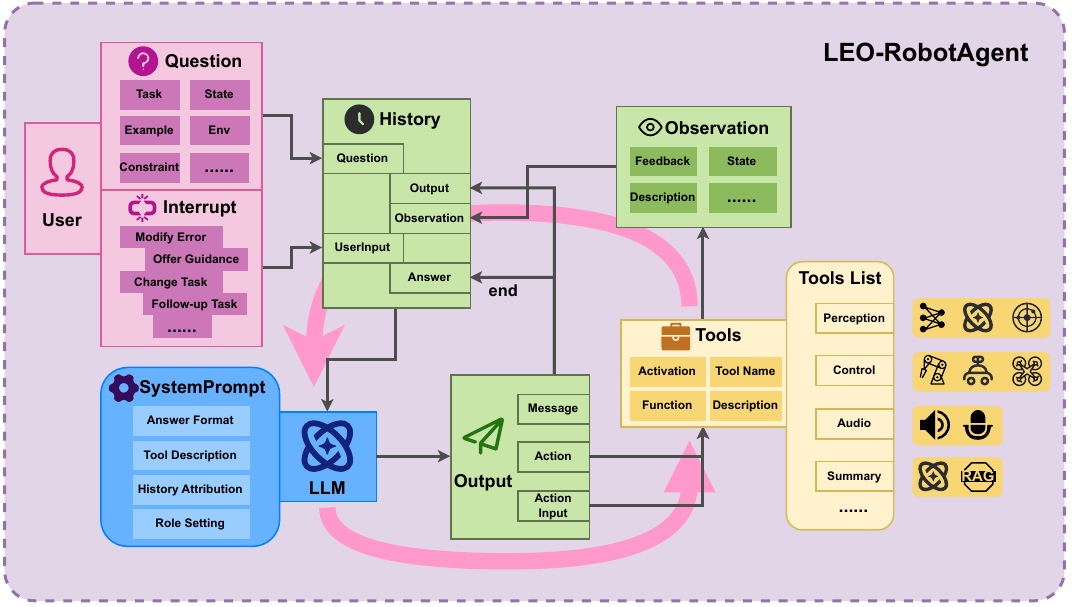}
\caption{Detailed implementation diagram of LEO-RobotAgent.}
\label{fig_2}
\vspace{-0.4cm}
\end{figure*}

The basic working principle of the LEO-RobotAgent framework is illustrated in Fig.~\ref{fig_1}. After users input task descriptions and relevant information, LLMs generate reasoning and planning content based on pre-defined prompts, while simultaneously invoking tools to execute actions. Each action yields corresponding observations, forming a closed feedback loop. During the whole process, users can interact with the LLMs at any time according to the feedback, to participate in and alter the task trajectory. The framework features a highly streamlined structure, functioning as a self-cycling engine that endows LLMs with sufficient autonomy and flexibility. This section will elaborate on the specific technical details and detailed system architecture. Fig.~\ref{fig_2} illustrates the high degree of flexibility and operational scalability of this streamlined framework through its specific implementation details, while the complete system design is presented in Fig.~\ref{fig_3}.

\subsection{Agent for Robotics}

\textit{1) LLM Configuration:} First, it should be clarified that the natural language text normally output by LLMs cannot be directly used for interaction; thus, we strictly constrain it to output content in JSON format. The pre-configured system prompt for LLMs includes key information such as requirements for JSON output content, tool-related details, interpretation of historical data, and the current role definition of LLMs, ensuring that LLMs can take correct actions according to task requirements and progress. The JSON output by LLMs must contain at least three components: Message, Action, and Action Input. Among them, the Message optionally conveys LLMs’ task planning, assessment of the current situation, and reasoning processes, while the other two components specify the action to be executed by LLMs in the current step and its corresponding input parameters. It is worth noting that LLMs must be required to output the Message first to ensure that the subsequent action is guided by the current reasoning.

\textit{2) Toolset Module:} The toolset encompasses all tools that can be invoked by LLMs. We can implement tools tailored to task requirements and register them in this module. Various core capabilities of robots, such as basic control, perception, and environmental interaction, can be realized in the form of tools; this module can even invoke other LLMs to construct a multi-agent architecture. Each tool entry must include at least the tool name and its corresponding function name, tool description (specifying input and output formats), and availability status of the tool for the current task. This module endows the framework with flexibility in behavioral capabilities and provides a simple yet convenient interface for users to configure tools according to practical demands. The output of a tool, namely the Observation, varies depending on the tool’s nature. It can be a description of target detection results, feedback on a UAV’s arrival at a waypoint, confirmation of successful robotic arm motion planning, analysis of received tasks, and so on.

\textit{3) Cycle Structure and History Mechanism:} With LLMs and the toolset in place, a basic agent loop can be constructed, where LLMs continuously reason and invoke tools, then proceed to the next step upon receiving feedback. During this process, historical records are gradually accumulated, which include user tasks, LLMs’ outputs at each step, tool observations from each iteration, and interim inputs from users. When LLMs determine that the task has been completed or cannot proceed further during execution, the loop will terminate and generate a final response. This response can be a task completion report or a description of the specific interruption cause.

\textit{4) Human-Robot Interaction Mechanism:} Task instructions input by users must at least specify the task content, essential initial states, and scenario descriptions. When necessary, practical examples and precautions can be provided to standardize the action logic of LLMs. If the current task permits human intervention and collaboration, users can interrupt task execution while the agent framework is running to correct the framework’s existing errors and provide guidance, temporarily modify task content, issue instructions for the next task phase, and so on. Based on this mechanism, the framework, as an intelligent agent, is endowed with complete two-way human-robot interaction capabilities.

\subsection{Complete Agent System}

\begin{figure}[!t]
\centering
\includegraphics[width=3.5in]{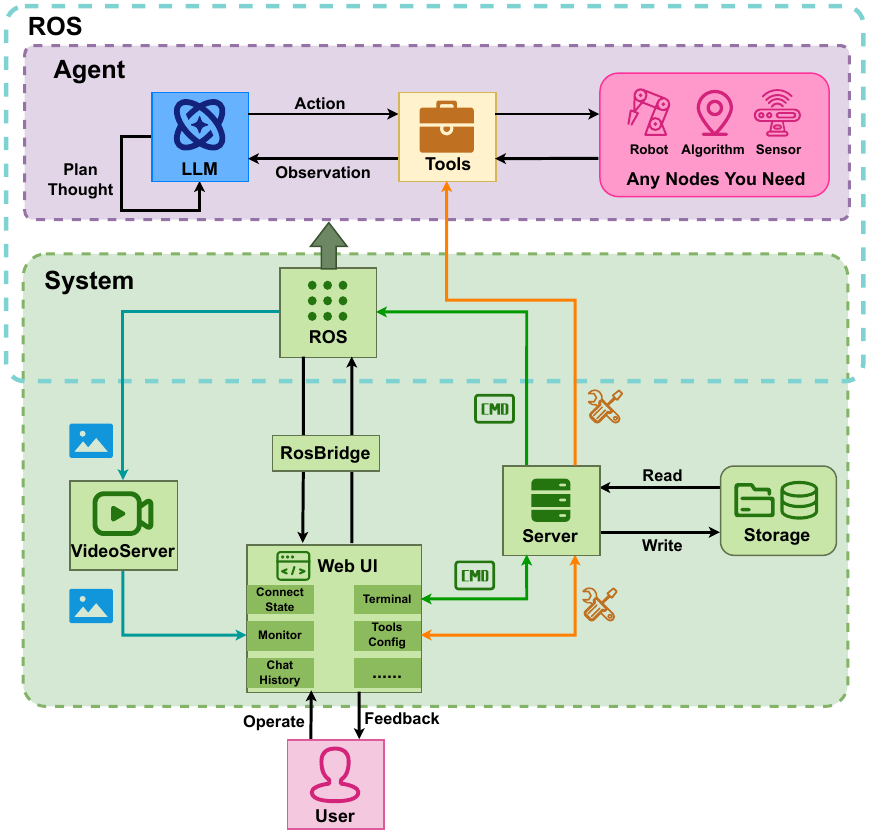}
\caption{An application system designed around LEO-RobotAgent.}
\label{fig_3}
\vspace{-0.6cm}
\end{figure}

As illustrated in Fig.~\ref{fig_3}, We have designed a complete, interactive ROS-based system around the framework elaborated in the previous section, where LLMs and the toolset are deeply integrated into the ROS system as Agent nodes. The tool module can effectively interface with a variety of tools; for instance, robot control nodes, visual perception nodes, Retrieval-Augmented Generation (RAG), simulation environments, and others can be registered in the toolset and invoked by the Agent node after their functions are implemented.

Dialogue messages, robot control, perception functions, and even tool feedback of the framework all rely on the topic mechanism of ROS for stable and long-term communication. For users, interaction with the ROS system is enabled via a visual interface we built based on Web applications, making the dialog and communication with the Agent framework no different from interactions on conventional LLM platforms. Specifically, topics are transmitted via RosBridge, and video streams are transmitted and displayed through a VideoServer.

In addition, we have refined tool registration, node startup/shutdown, task preconfiguration, and other operations based on WebSocket, such that debugging and operation of the entire system can be conducted almost entirely within this application interface. This architecture provides a paradigm for constructing LLM-agent applications in the robotics domain.

\section{Experiments}
This section validates the feasibility, general value and efficiency of the LEO-RobotAgent framework through a series of experiments. The experimental tasks cover multiple types of robots, tasks with varying complexity levels, and their corresponding scenarios. Additionally, comparative experiments are conducted to demonstrate how to interact with the agent using different prompt techniques, as well as to illustrate the characteristics scenarios of agent frameworks with different structures and underlying principles. The Qwen3-Max large language model was adopted for all experiments in this study.

\subsection{Feasibility Verification and Real UAV Experiment}

\begin{figure*}[!t]
\centering
\includegraphics[width=6.8in]{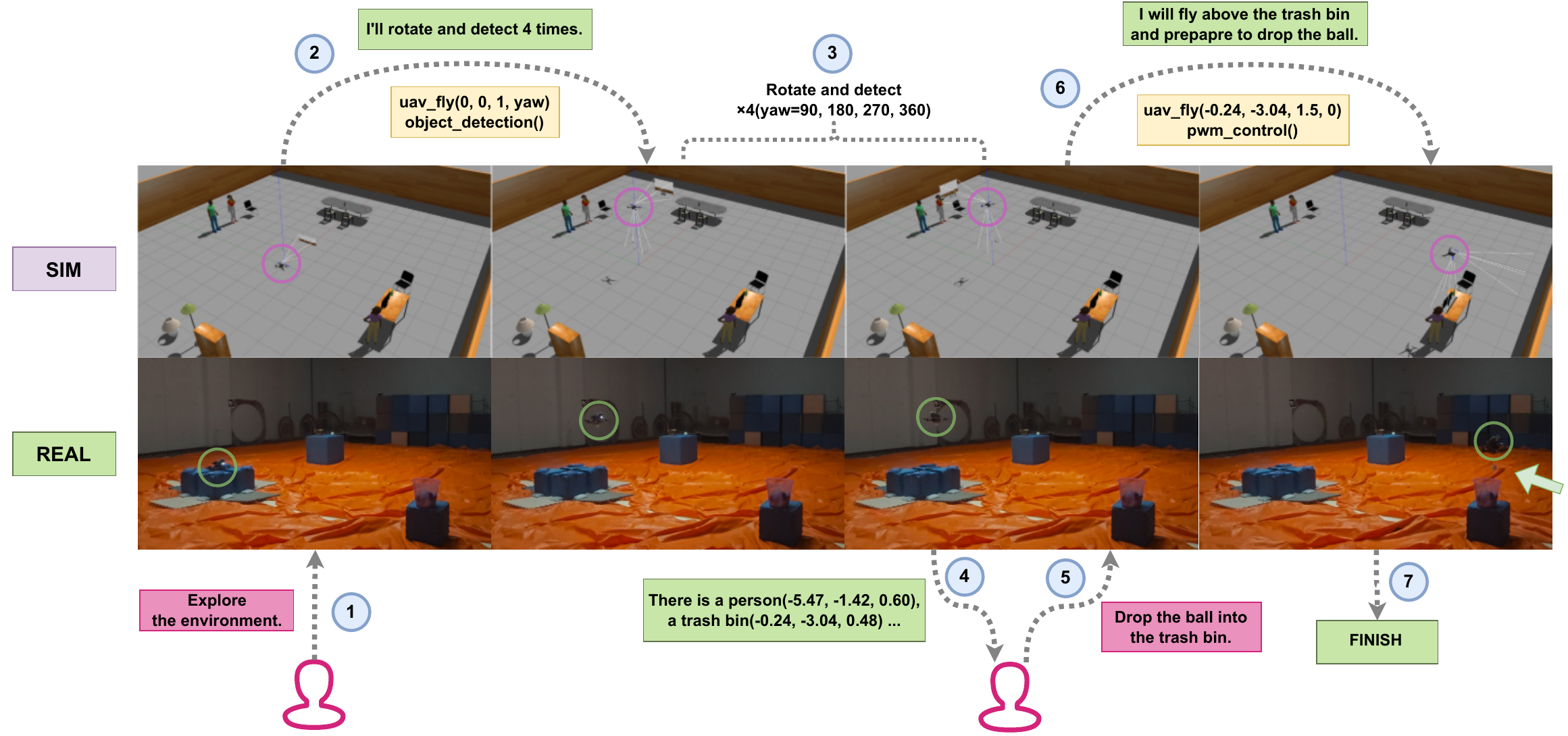}
\caption{Feasibility verification and real experiment.}
\label{expA}
\end{figure*}

This subsection conducts simulation experiments on a UAV via a moderately complex object-search task to initially verify that the framework can autonomously complete tasks and that its human-robot interaction function is practical. The same task is further executed on a real UAV to validate the algorithm’s easy sim-to-real transfer. The UAV is chosen because it integrates control, perception, and interaction capabilities, representing the general competencies of robots.

Equipped with a depth camera, the UAV is required to search for a specified nearby object in a small scenario using target detection, and then fly above it after receiving a secondary user instruction to complete the task. This task employs the UAV’s position-hold flight control and target detection as core tools. In real experiments, an electromagnetically attached iron ball is mounted under the UAV, which must additionally drop the ball into a target container.

The experimental results are illustrated in Fig.~\ref{expA}. Ten trials were conducted for both simulation and real-world experiments, with 9 and 7 successful attempts respectively. The success rates were high and close to each other, demonstrating good algorithm stability. Notably, partial failures in real-world experiments were attributed to limitations in the accuracy of flight control tools and object localization errors, which prevented the UAV from precisely hovering above the trash can.

This experiment verifies that the framework can be effectively deployed on UAVs, and the algorithm enables straightforward sim-to-real transfer, demonstrating feasibility for engineering implementation. Since the reasoning logic of LLMs itself is not affected by whether the task is executed in a virtual or physical environment, the sim-to-real gap of this framework largely depends on the sim-to-real gaps of the robot and the tools we have implemented.

\subsection{Prompt Experiment}

\begin{figure}[!t]
\centering
\subfloat[Room]{\includegraphics[width=0.235\textwidth]{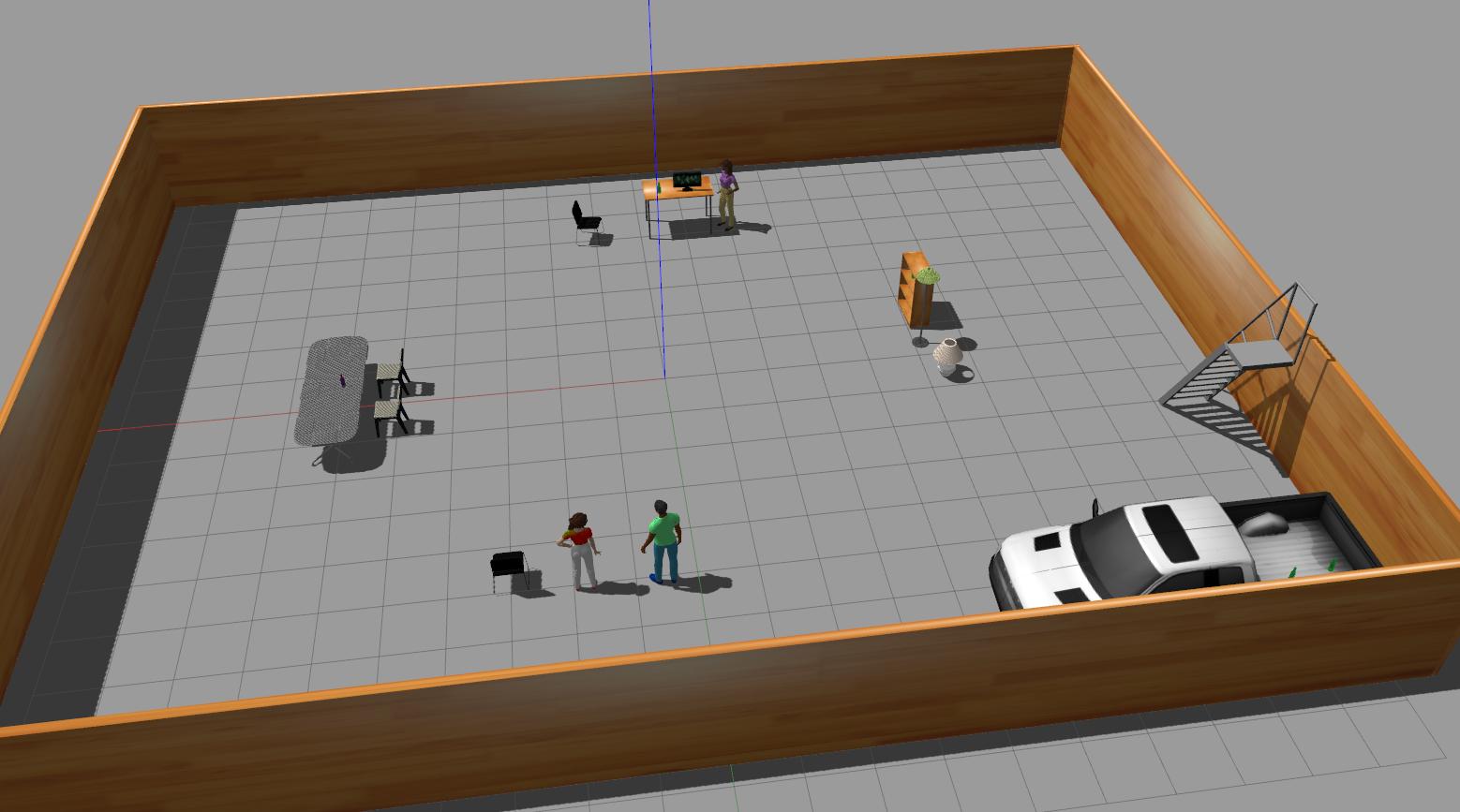}%
\label{fig_first_case}}
\hfil
\subfloat[City]{\includegraphics[width=0.235\textwidth]{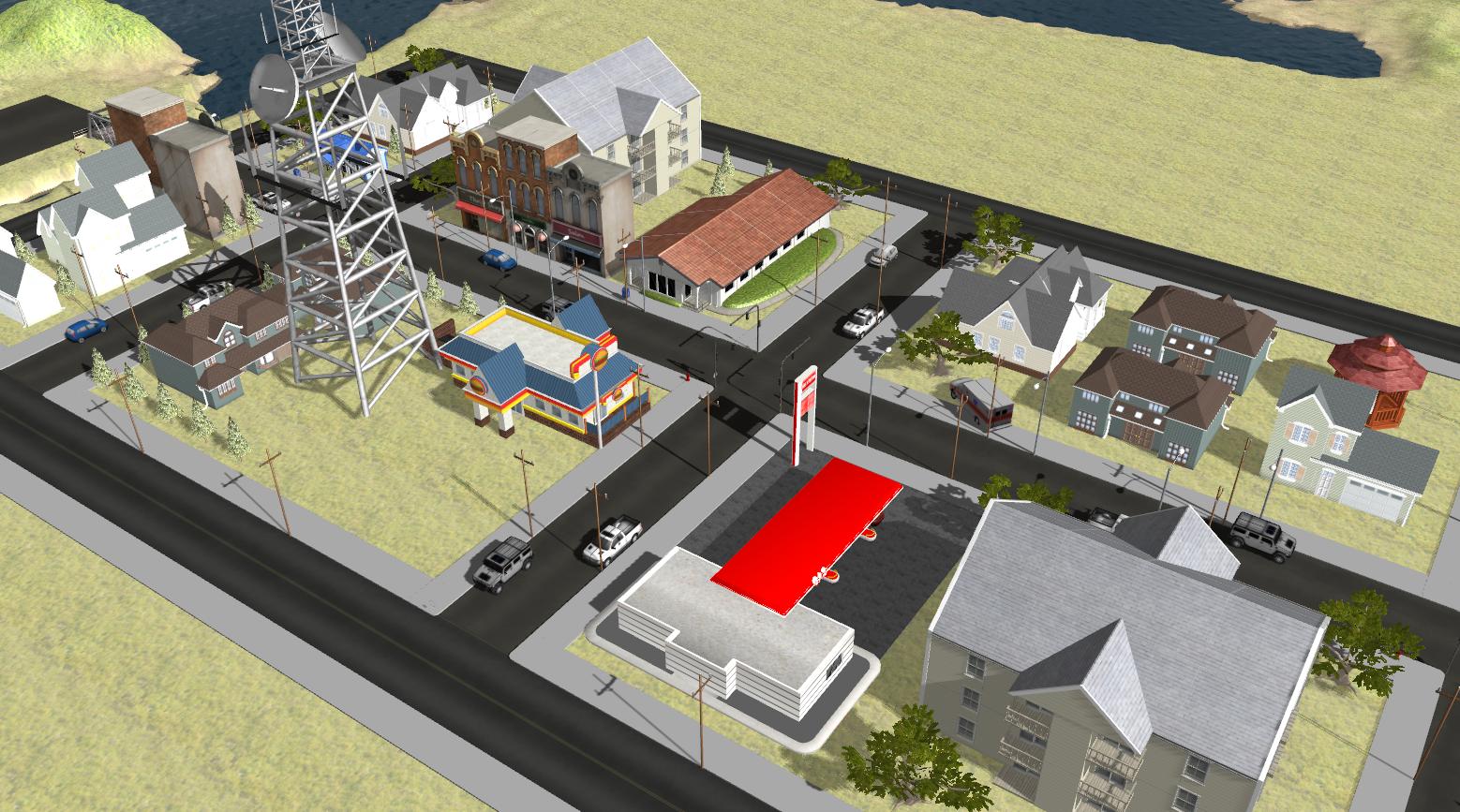}%
\label{fig_second_case}}
\caption{UAV conducting indoor and urban searching tasks with prompt
engineering techniques.}
\label{expB-1}
\vspace{-0.6cm}
\end{figure}

\begin{table*}[!t]
\caption{Results of Prompt Experiment in Two Scenarios}
\label{tab:expB}
\centering
\renewcommand{\arraystretch}{1} 

\begin{tabular}{lcccccccc}
\toprule
\multirow{2}{*}{Method} &
\multicolumn{4}{c}{Room Scenario} &
\multicolumn{4}{c}{City Scenario} \\
\cmidrule(lr){2-5} \cmidrule(lr){6-9}
& score(20) & token usage & time(s) & time/item(s) 
& success rate(\%) & token usage & time(s) & success time(s) \\
\midrule

zero-shot
& 7.13±4.28  & 6714  & 59.83  & 8.39 
& 20.0   & 32656 & 175.23 & 183.83 \\

one-shot
& 17.33±2.18 & 18537 & 101.93 & 5.88
& 50.0   & 32048 & 156.65 & 123.38 \\

CoT
& 16.50±2.69 & 39937 & 150.14 & 9.10
& 60.0   & 37791 & 155.00 & 126.93 \\

one-shot+CoT
& \textbf{17.80±1.26} & 36476 & 133.23 & 7.48
& \textbf{70.0}   & 44985 & 180.22 & 172.81 \\

\bottomrule
\end{tabular}
\vspace{-0.4cm}
\end{table*}


\begin{figure*}[!t]
\centering
\subfloat[Room]{\includegraphics[width=0.49\textwidth]{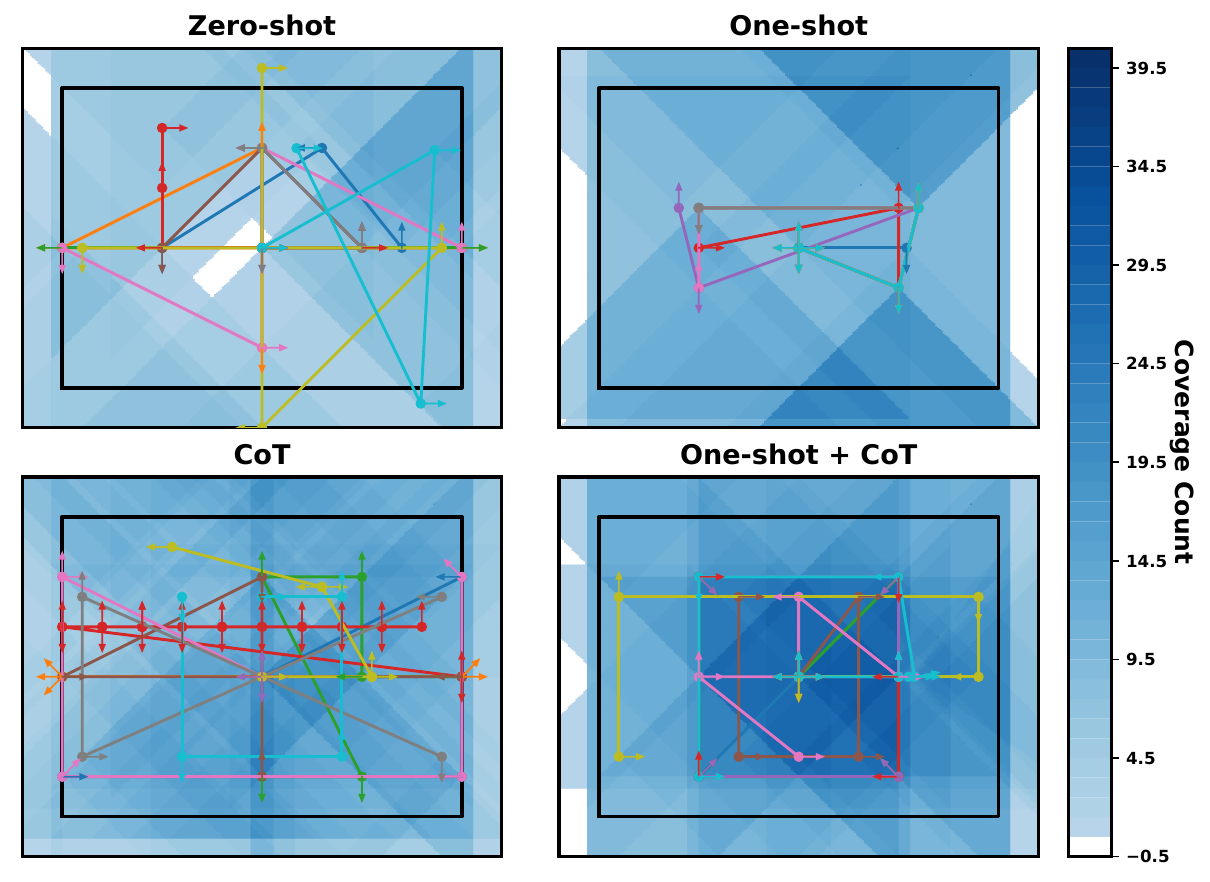}%
\label{path_room}}
\hfil
\subfloat[City]{\includegraphics[width=0.49\textwidth]{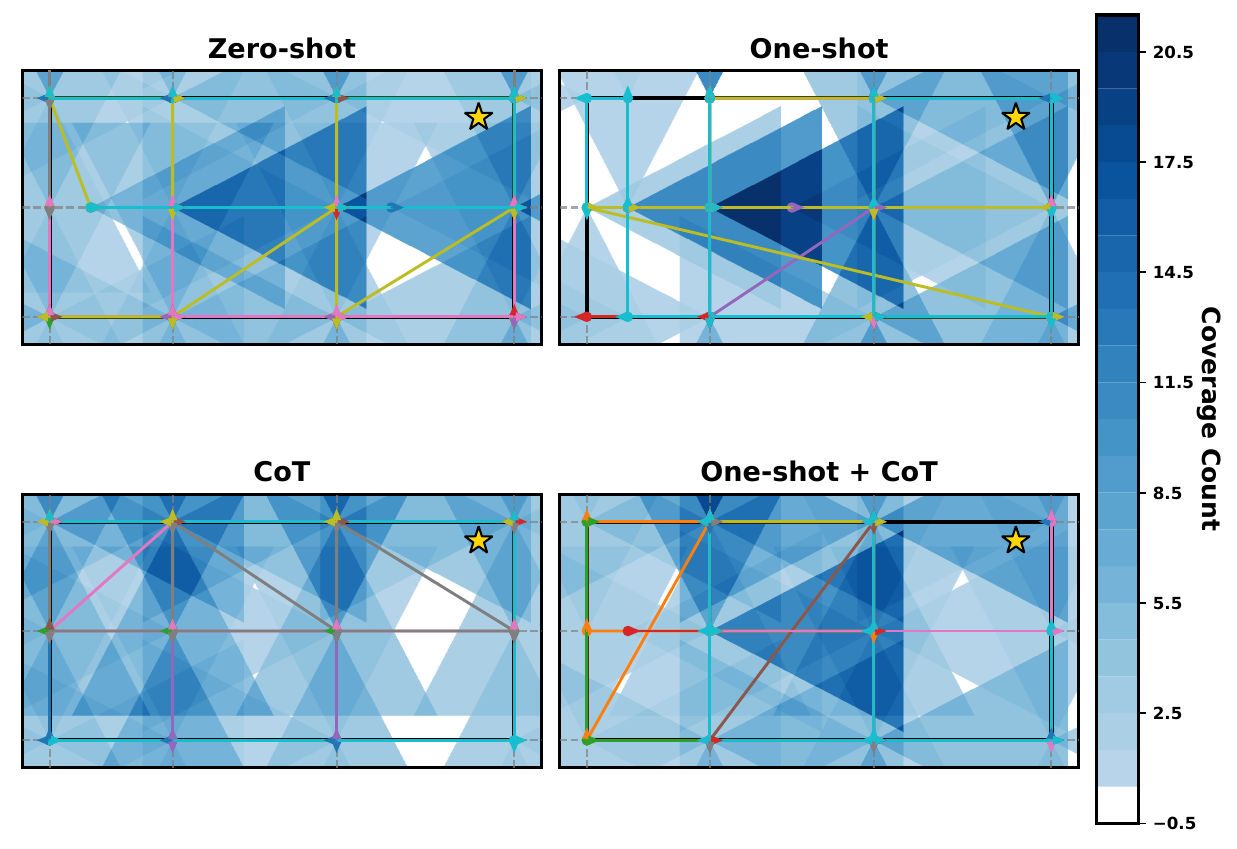}%
\label{path_city}}
\caption{Field of view coverage map for UAV searching tasks in indoor and urban scenarios.}
\label{expB-2}
\end{figure*}

The design of prompts during task deployment largely determines the output quality of LLMs, which in turn affects task execution performance. This experiment verifies the impact of Chain-of-Thought (CoT)\cite{cot} and one-shot\cite{gpt3} on the task planning effectiveness and performance of the framework. CoT requires prompts to guide the agent in reasoning or provide reasoning processes, while one-shot entails providing an example to assist LLMs in comprehension.

The task is divided into two subtasks: indoor small-scenario search and urban large-scenario search (Fig.~\ref{expB-1}). The former requires the UAV to use target detection to locate all identifiable objects indoors as much as possible, while the latter only requires finding a target building (pavilion) in a large urban scene using a VLM as the perception tool.

Experimental results are presented in Table~\ref{tab:expB}, where each method was tested 10 times per scenario to obtain average outcomes. The time/item refers to the time required for the UAV to locate each unique object, while success time denotes the total duration of the task when it is completed successfully. It is evident that both methods effectively improve task planning performance, with the best results achieved when combining them. One-shot yields the fastest task execution in successful cases, as it enables relatively robust operations based on existing examples, ensuring a high performance floor. In contrast, CoT incurs substantial token and time overhead, since it requires LLMs to conduct extensive, step-by-step reasoning and planning.

As visualized in Fig.~\ref{expB-2} for the UAV’s field-of-view coverage during searching tasks, thick black rectangles denote scenario boundaries. Stars denote the target to find. Deeper blue indicates a higher number of times the UAV’s field of view has covered the location. Different colored paths represent the search process of each task, marking all waypoints and yaw angles. Without prompts, the UAV often performs ineffective searches (e.g., reaching boundaries and pointing the camera outward). One-shot leads to exploration patterns that closely mimic the provided example but fails to effectively cover corner details. CoT fosters more divergent reasoning in LLMs, generating feasible and innovative search paths. The combined approach produces search paths that are both thorough and efficient, and similar trends are observed in urban search scenarios.

This section’s experiments confirm that common prompting techniques in the LLM domain can be effectively applied to robotic agent frameworks. Therefore, when assigning tasks, we should provide as much necessary information as possible if conditions permit, such as pre-emptive avoidance of potential errors, demonstrations of feasible planning strategies, and logical analyses of reasoning processes.

\subsection{Agent Framework Architecture Comparison Experiment}

\begin{figure*}[!t]
\centering
\includegraphics[width=6.6in]{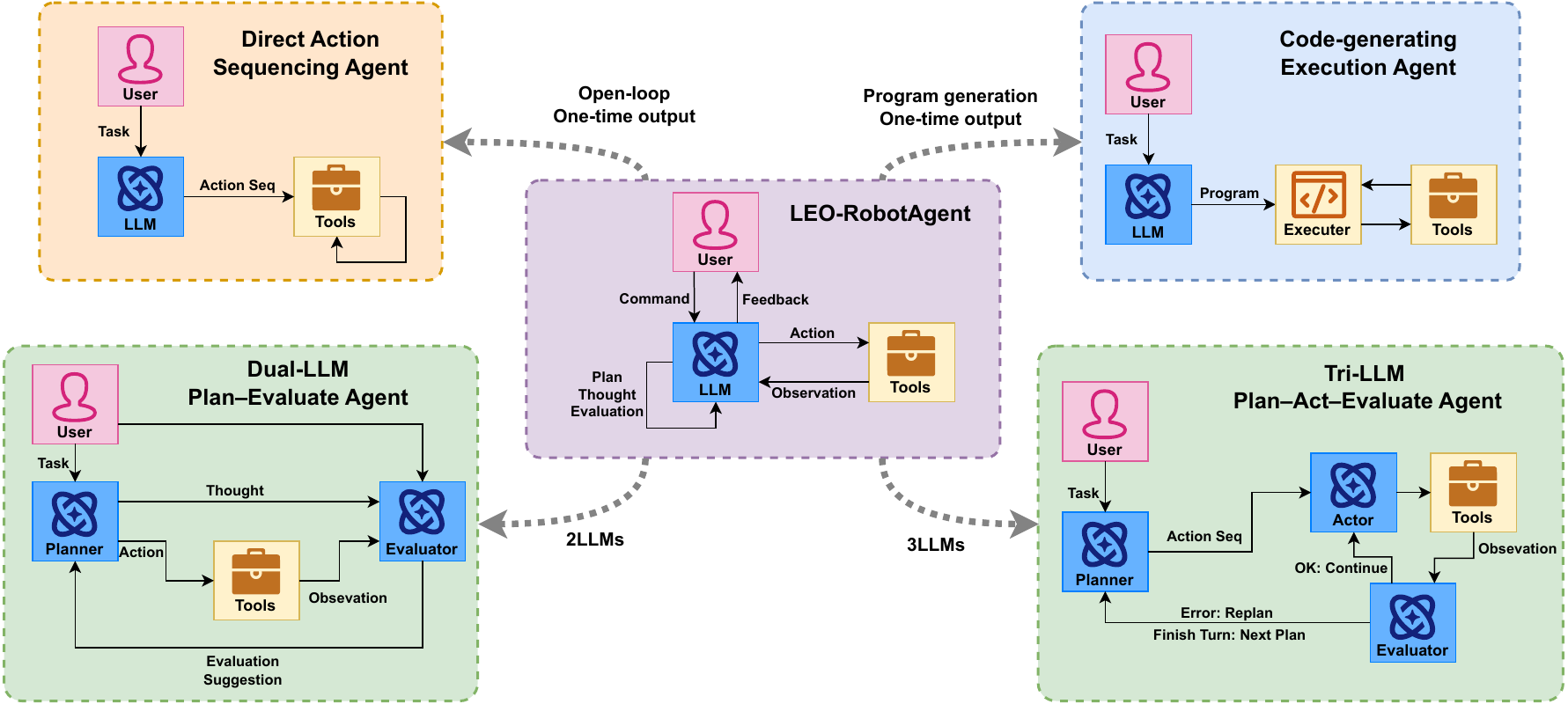}
\caption{LEO-RobotAgent and four other agent schemes.}
\label{expC-1}
\vspace{-0.4cm}
\end{figure*}

\begin{figure}[!t]
\centering
\subfloat[Robot]{\includegraphics[width=0.235\textwidth]{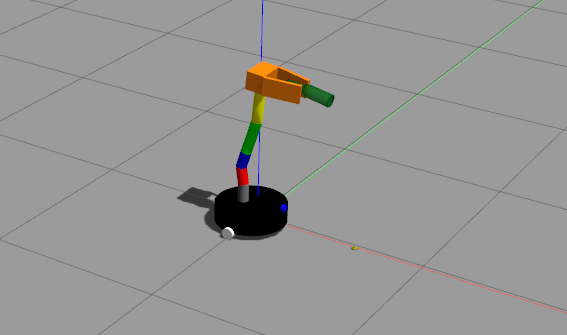}%
\label{fig_first_case}}
\hfill
\subfloat[Cafe]{\includegraphics[width=0.235\textwidth]{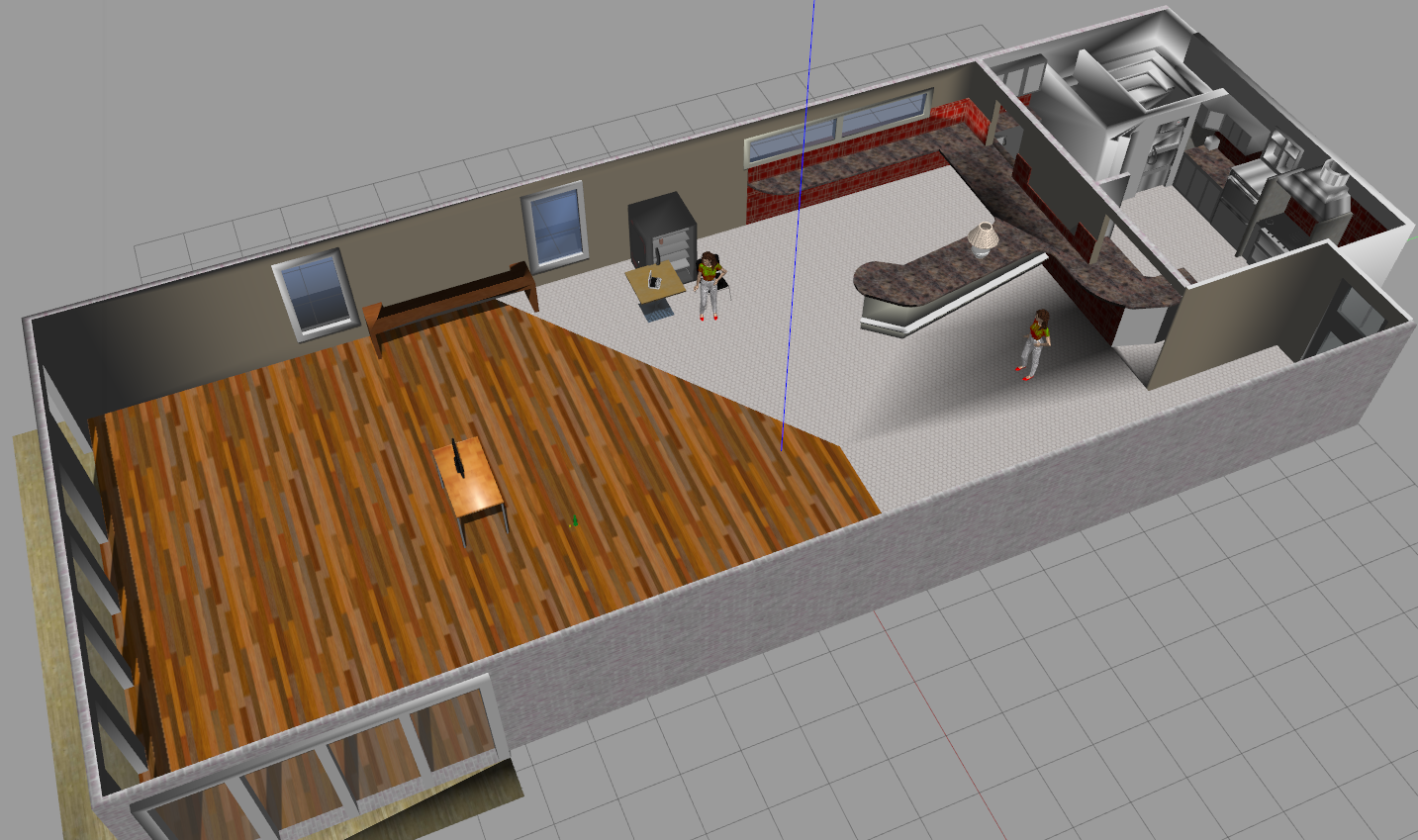}%
\label{fig_second_case}}
\caption{The wheeled robot with robotic arm and the map of a cafe for Agent Framework Architecture Comparison Experiment.}
\label{expC-2}
\vspace{-0.5cm}
\end{figure}

\begin{table}[!t]
\setlength{\textfloatsep}{5pt plus 1.0pt minus 2.0pt}
\caption{Overview of Agent Comparison Experiment}
\label{tab:expC-1}
\centering
\renewcommand{\arraystretch}{1.0} 

\begin{tabularx}{\columnwidth}{@{} >{\raggedright\arraybackslash}p{1.2cm} >{\raggedright\arraybackslash}X @{\hspace{5pt}}c @{\hspace{5pt}}c @{}}
\toprule
\textbf{Task} & \textbf{Description} & \textbf{Perc.} & \textbf{NLU} \\
\midrule

Delivery & 
Given the coordinates of three bottles and three target spots, the robot must pick and place them in an arbitrary order and finally return to the origin. & 
$\times$ & 
$\times$ \\ 
\midrule

Searching & 
Command the robot to search for the nearest bottle in the scene, pick it up, and return to the origin. & 
\checkmark & 
$\times$ \\ 
\midrule

Handover & 
Two people are located by a chair and a lamp respectively. The robot is instructed to find the former, approach to receive a sub-task, and return to the origin after completion. 
\par\vspace{2pt}
Sub-task: Retrieve a bottle and place it near the other person. & 
\checkmark & 
\checkmark \\

\bottomrule
\end{tabularx}
\end{table}

This subsection compares the performance of agent schemes with different principles and architectures across tasks of varying complexity. Based on the core ideas from the outstanding prior works mentioned earlier, we summarize four other highly distinctive schemes(Fig.~\ref{expC-1}). Thanks to the simplicity and high scalability of our framework, these abstracted architectures can be easily adapted to the toolset, human-robot interaction mechanism, and application system of our framework. The schemes are described as follows:

\textit{1) Direct Action Sequencing Agent (DAS):} LLMs directly generate a complete sequence of actions based on the task text and available toolset, and the system executes the sequence step-by-step in a fully open-loop manner.

\textit{2) Code-generating Execution Agent (CGE):} LLMs output executable Python code, where the action statements still do not exceed the scope of the toolset, and the code is directly executed after review.

\textit{3) Dual-LLM Plan–Evaluate Agent (DLLMs):} This architecture relies on the collaboration of two LLMs: the Planner is responsible for planning, reasoning, and executing actions; the Evaluator evaluates the planning content and execution feedback and then puts forward suggestions for the former’s outputs. The process loops in this way and accumulates historical records.

\textit{4) Tri-LLM Plan–Act–Evaluate Agent (TLLMs):} This architecture decomposes agent functions into three LLMs: the Planner generates high-level reasoning and task plans; the Actor converts the plans into specific tool calls; the Evaluator analyzes execution performance based on observation results and provides new inputs. Each plan of the Planner operates in a large loop, which can divide the task into a single or multiple stages; within each plan, the Actor executes step-by-step and the Evaluator conducts gradual evaluation, which operates in a small loop.

Owing to the differing capability boundaries of these schemes: The DAS lacks a closed loop and cannot perform perception (Perc.); The CGE can achieve closed-loop functionality via program design but has no natural language understanding (NLU) capability. For this experiment, we designed three tasks of varying complexity to ensure that every scheme could participate in at least one task. The task platform was a wheeled mobile robot equipped with a robotic arm (Fig.~\ref{expC-2}); this robot structure was fully custom-designed in a simulation environment and could perform fixed-point movement, as well as simple grasping and releasing operations.

The three tasks are detailed in Table~\ref{tab:expC-1}. Task 1 only requires completing object delivery in a non-fixed sequential manner. Task 2 necessitates distinguishing different objects of the same category based on their relative distance relationships. In contrast, Task 3 involves fulfilling nested primary and sub-tasks, which requires understanding sub-tasks presented in natural language and conducting secondary planning; for this purpose, we introduce the summarization capability of LLMs as a new tool. Given the lengthy workflow of this task, it also tests the agent’s ability to plan and memorize long-horizon tasks.

\begin{table*}[!t]
\caption{Results of Agent Comparison Experiment}
\label{tab:expC-2}
\centering
\renewcommand{\arraystretch}{1.0} 

\begin{tabular}{c|c|ccccc}
\toprule
\textbf{Task} & \textbf{Agent} & \textbf{Score (10)} & \textbf{Token Usage} & \textbf{Time (s)} & \textbf{Success Time (s)} & \textbf{Perfect Rate (\%)} \\
\midrule

\multirow{5}{*}{Delivery} 
& DAS   & 9.156±1.296 & 1054  & 232.18 & 228.14 & 68.75 \\
& CGE   & \textbf{9.344±1.182} & 968 & 208.89 & 207.87 & \textbf{75.00} \\
& DLLMs & 7.938±1.550 & 24684 & 233.35 & 245.53 & 12.50 \\
& TLLMs & 8.375±1.763 & 43278 & 277.37 & 302.32 & 37.50 \\
& LEO (ours) & 9.156±1.221 & 19000 & 264.62 & 272.71 & 62.50 \\
& LEO (human) & 9.786±1.053 & 20378 & 272.85 & 288.41 & 81.25 \\
\midrule 

\multirow{5}{*}{Searching} 
& DAS   & -- & -- & -- & -- & -- \\
& CGE   & 5.375±1.717 & 1447 & 73.76 & 97.93 & 43.75 \\
& DLLMs & 3.125±2.672 & 54056 & 203.77 & 165.78 & 18.75 \\
& TLLMs & 5.875±3.501 & 115848 & 266.84 & 190.32 & 43.75 \\
& LEO (ours) & \textbf{7.875±0.866} & 14597 & 141.44 & 120.08 & \textbf{56.25} \\
& LEO (human) & 8.661±0.724 & 16827 & 160.58 & 138.59 & 75.00 \\
\midrule 

\multirow{5}{*}{Handover} 
& DAS   & -- & -- & -- & -- & -- \\
& CGE   & -- & -- & -- & -- & -- \\
& DLLMs & 4.933±2.514 & 58757 & 313.23 & 276.82 & 12.50 \\
& TLLMs & 4.867±2.951 & 188596 & 432.44 & 597.95 & 12.50 \\
& LEO (ours) & \textbf{7.867±1.092} & 33577 & 257.71 & 265.61 & \textbf{43.75} \\
& LEO (human) & 8.537±0.852 & 45785 & 279.71 & 282.10 & 62.50 \\

\bottomrule
\end{tabular}
\vspace{-0.4cm}
\end{table*}

We conducted 16 experiments for each scheme in every scenario, using the same large language model and identical task descriptions; the results are presented in Table~\ref{tab:expC-2}. The score was designed based on the degree of task completion, with a full score of 10 points. Perfect Rate denotes the ratio of achieving a full score, i.e., completing the task in its entirety.

It can be observed that for Task 1, which is simple and well-defined, one-time generation methods such as DAS and CGE exhibit strong stability. Moreover, these methods incur extremely low token and time overhead in the tasks they can execute, indicating that they remain suitable for relatively simple and clearly specified tasks.

In contrast, DLLMs and TLLMs—frameworks with multi-LLM division of labor—unexpectedly yielded low scores. This is largely because multiple roles in the framework require tuning with corresponding preset prompts, and despite investing far more effort in this calibration than for other agent schemes, this was the best performance achieved. Coordination among multiple LLMs leads to a rapid surge in token consumption and a higher propensity for hallucinations, which hinders the execution of normal plans or results in the omission of key details. For instance, the Evaluator often incorrectly triggers replanning operations, thereby impeding the normal execution of plans; alternatively, misinterpretation of overly lengthy text may result in the omission of detailed steps such as returning to the origin, which are specified in the task requirements.

Our framework also delivers promising performance on Task 1, where DAS and CGE excel. For the more challenging Tasks 2 and 3, it incurs significantly lower time and token costs. Furthermore, our framework achieves consistently high scores with low standard deviations across three tasks. This superior and stable performance can be attributed to its advantages over the more complex DLLMs and TLLMs: it generates far fewer hallucinations and stably retains long-term memory throughout the experimental process. Furthermore, to quantitatively evaluate the human-in-the-loop mechanism, we introduce the LEO (human) setting, where the user is allowed to intervene exactly once when the agent encounters an error. While this significantly improves performance, it also introduces extra burdens, explicitly reflected in increased Token Usage and execution Time.

This experiment indicates that LEO-RobotAgent, leveraging its more streamlined agent architecture, achieves lower debugging overhead and superior task planning performance with enhanced robustness, which validates the principle of ``less is more". Meanwhile, we successfully completed the experiment using this fully customized robot architecture, which also verified the universal value of the framework.

\section{Conclusion}

This paper proposes LEO-RobotAgent, a streamlined, general-purpose agent framework tailored for robotic embodied operators. It exhibits remarkable versatility across multiple dimensions, including robot types, task scenarios, and task contents, with a high degree of autonomy. A complete application-level system built around LEO-RobotAgent significantly reduces the difficulty of human-robot interaction and mutual understanding. Moreover, the full solution features straightforward sim-to-real transfer and strong portability. It is worth noting that rather than proposing fundamentally novel foundational learning algorithms, the core contribution of this work lies in the system-level integration, robust architectural design, and practical deployment across heterogeneous robotic platforms.

The prompt experiment shows that imparting sufficient reasoning guidance and task-related information such as a reference example to the agent yields significant improvements in task planning performance. Through agent comparison experiments, we conclude that a streamlined framework can better unleash the capabilities of LLMs and reduce the debugging burden for users. Our framework assigns multiple responsibilities to a single LLM and allows it to sequentially complete the reasoning and action stages within each step, thereby endowing the agent with a more coherent and unified thinking process and a rational planning path. Additionally, the adoption of auxiliary LLMs or VLMs as tools for summarization and perception in our experiments demonstrates that multi-role architectures can be unified via our framework’s toolset module.

The experiments also reveal the limitations of our method. Given the weak spatial common-sense understanding of current LLMs, it is necessary to guide them with processing strategies for three-dimensional (3D) space to enable the agent to execute correct control operations (e.g., adjusting angles to face objects directly). Thus, enhancing LLMs’ real-world spatial cognition is a critical direction for advancing robotic agent frameworks, and also a key topic for realizing artificial general intelligence (AGI) in the future. This work is a preliminary exploration, and we expect more outstanding research to deliver further progress in this field.

\bibliographystyle{IEEEtran}
\bibliography{IEEEabrv, references}

\begin{thebibliography}{10}
\providecommand{\url}[1]{#1}
\csname url@samestyle\endcsname
\providecommand{\newblock}{\relax}
\providecommand{\bibinfo}[2]{#2}
\providecommand{\BIBentrySTDinterwordspacing}{\spaceskip=0pt\relax}
\providecommand{\BIBentryALTinterwordstretchfactor}{4}
\providecommand{\BIBentryALTinterwordspacing}{\spaceskip=\fontdimen2\font plus
\BIBentryALTinterwordstretchfactor\fontdimen3\font minus
  \fontdimen4\font\relax}
\providecommand{\BIBforeignlanguage}[2]{{%
\expandafter\ifx\csname l@#1\endcsname\relax
\typeout{** WARNING: IEEEtran.bst: No hyphenation pattern has been}%
\typeout{** loaded for the language `#1'. Using the pattern for}%
\typeout{** the default language instead.}%
\else
\language=\csname l@#1\endcsname
\fi
#2}}
\providecommand{\BIBdecl}{\relax}
\BIBdecl

\bibitem{uavs}
Y.~Tian, F.~Lin, Y.~Li, T.~Zhang, Q.~Zhang, X.~Fu, J.~Huang, X.~Dai, Y.~Wang,
  C.~Tian \emph{et~al.}, ``Uavs meet llms: Overviews and perspectives towards
  agentic low-altitude mobility,'' \emph{Information Fusion}, vol. 122, p.
  103158, 2025.

\bibitem{robots}
F.~Zeng, W.~Gan, Y.~Wang, N.~Liu, and P.~S. Yu, ``Large language models for
  robotics: A survey,'' \emph{arXiv preprint arXiv:2311.07226}, 2023.

\bibitem{react}
S.~Yao, J.~Zhao, D.~Yu, N.~Du, I.~Shafran, K.~R. Narasimhan, and Y.~Cao,
  ``React: Synergizing reasoning and acting in language models,'' in \emph{The
  eleventh international conference on learning representations}, 2022.

\bibitem{assist1}
X.~Zhao, M.~Li, C.~Weber, M.~B. Hafez, and S.~Wermter, ``Chat with the
  environment: Interactive multimodal perception using large language models,''
  in \emph{2023 IEEE/RSJ International Conference on Intelligent Robots and
  Systems (IROS)}.\hskip 1em plus 0.5em minus 0.4em\relax IEEE, 2023, pp.
  3590--3596.

\bibitem{assist2}
Y.~Ding, X.~Zhang, C.~Paxton, and S.~Zhang, ``Task and motion planning with
  large language models for object rearrangement,'' in \emph{2023 IEEE/RSJ
  International Conference on Intelligent Robots and Systems (IROS)}.\hskip 1em
  plus 0.5em minus 0.4em\relax IEEE, 2023, pp. 2086--2092.

\bibitem{assist3}
A.~Tagliabue, K.~Kondo, T.~Zhao, M.~Peterson, C.~T. Tewari, and J.~P. How,
  ``Real: Resilience and adaptation using large language models on autonomous
  aerial robots,'' in \emph{2024 IEEE 63rd Conference on Decision and Control
  (CDC)}.\hskip 1em plus 0.5em minus 0.4em\relax IEEE, 2024, pp. 1539--1546.

\bibitem{assist4}
V.~G. Goecks and N.~R. Waytowich, ``Disasterresponsegpt: Large language models
  for accelerated plan of action development in disaster response scenarios,''
  \emph{arXiv preprint arXiv:2306.17271}, 2023.

\bibitem{assist5}
Z.~Yuan, F.~Xie, and T.~Ji, ``Patrol agent: An autonomous uav framework for
  urban patrol using on board vision language model and on cloud large language
  model,'' in \emph{2024 6th International Conference on Robotics and Computer
  Vision (ICRCV)}.\hskip 1em plus 0.5em minus 0.4em\relax IEEE, 2024, pp.
  237--242.

\bibitem{codegeneration1}
I.~Singh, V.~Blukis, A.~Mousavian, A.~Goyal, D.~Xu, J.~Tremblay, D.~Fox,
  J.~Thomason, and A.~Garg, ``Progprompt: Generating situated robot task plans
  using large language models,'' \emph{arXiv preprint arXiv:2209.11302}, 2022.

\bibitem{codegeneration2}
G.~Chen, X.~Yu, N.~Ling, and L.~Zhong, ``Typefly: Flying drones with large
  language model,'' \emph{arXiv preprint arXiv:2312.14950}, 2023.

\bibitem{codegeneration3}
Z.~Hu, F.~Lucchetti, C.~Schlesinger, Y.~Saxena, A.~Freeman, S.~Modak, A.~Guha,
  and J.~Biswas, ``Deploying and evaluating llms to program service mobile
  robots,'' \emph{IEEE Robotics and Automation Letters}, vol.~9, no.~3, pp.
  2853--2860, 2024.

\bibitem{vla1}
Y.~Yin, Z.~Wang, Y.~Sharma, D.~Niu, T.~Darrell, and R.~Herzig, ``In-context
  learning enables robot action prediction in llms,'' in \emph{2025 IEEE
  International Conference on Robotics and Automation (ICRA)}.\hskip 1em plus
  0.5em minus 0.4em\relax IEEE, 2025, pp. 8972--8979.

\bibitem{vla2}
R.~Mon-Williams, G.~Li, R.~Long, W.~Du, and C.~G. Lucas, ``Embodied large
  language models enable robots to complete complex tasks in unpredictable
  environments,'' \emph{Nature Machine Intelligence}, pp. 1--10, 2025.

\bibitem{agent1}
H.~Zhao, F.~Pan, H.~Ping, and Y.~Zhou, ``Agent as cerebrum, controller as
  cerebellum: Implementing an embodied lmm-based agent on drones,'' \emph{arXiv
  preprint arXiv:2311.15033}, 2023.

\bibitem{agent2}
L.~Sun, D.~K. Jha, C.~Hori, S.~Jain, R.~Corcodel, X.~Zhu, M.~Tomizuka, and
  D.~Romeres, ``Interactive planning using large language models for partially
  observable robotic tasks,'' in \emph{2024 IEEE International Conference on
  Robotics and Automation (ICRA)}.\hskip 1em plus 0.5em minus 0.4em\relax IEEE,
  2024, pp. 14\,054--14\,061.

\bibitem{agent3}
H.~Singh, R.~J. Das, M.~Han, P.~Nakov, and I.~Laptev, ``Malmm: Multi-agent
  large language models for zero-shot robotics manipulation,'' \emph{arXiv
  preprint arXiv:2411.17636}, 2024.

\bibitem{agent4}
F.~Joublin, A.~Ceravola, P.~Smirnov, F.~Ocker, J.~Deigmoeller, A.~Belardinelli,
  C.~Wang, S.~Hasler, D.~Tanneberg, and M.~Gienger, ``Copal: corrective
  planning of robot actions with large language models,'' in \emph{2024 ieee
  international conference on robotics and automation (ICRA)}.\hskip 1em plus
  0.5em minus 0.4em\relax IEEE, 2024, pp. 8664--8670.

\bibitem{other1}
D.~Driess, F.~Xia, M.~S. Sajjadi, C.~Lynch, A.~Chowdhery, A.~Wahid, J.~Tompson,
  Q.~Vuong, T.~Yu, W.~Huang \emph{et~al.}, ``Palm-e: An embodied multimodal
  language model,'' 2023.

\bibitem{other2}
S.~Zhang, Z.~Xu, P.~Liu, X.~Yu, Y.~Li, Q.~Gao, Z.~Fei, Z.~Yin, Z.~Wu, Y.-G.
  Jiang \emph{et~al.}, ``Vlabench: A large-scale benchmark for
  language-conditioned robotics manipulation with long-horizon reasoning
  tasks,'' in \emph{Proceedings of the IEEE/CVF International Conference on
  Computer Vision}, 2025, pp. 11\,142--11\,152.

\bibitem{qwen2.5}
B.~Hui, J.~Yang, Z.~Cui, J.~Yang, D.~Liu, L.~Zhang, T.~Liu, J.~Zhang, B.~Yu,
  K.~Lu \emph{et~al.}, ``Qwen2. 5-coder technical report,'' \emph{arXiv
  preprint arXiv:2409.12186}, 2024.

\bibitem{qwen3}
A.~Yang, A.~Li, B.~Yang, B.~Zhang, B.~Hui, B.~Zheng, B.~Yu, C.~Gao, C.~Huang,
  C.~Lv \emph{et~al.}, ``Qwen3 technical report,'' \emph{arXiv preprint
  arXiv:2505.09388}, 2025.

\bibitem{other3}
J.~Zhao and X.~Lin, ``General-purpose aerial intelligent agents empowered by
  large language models,'' \emph{arXiv preprint arXiv:2503.08302}, 2025.

\bibitem{other4}
M.~Zhang, Q.~Dai, Y.~Yang, J.~Bao, D.~Chen, K.~Qiu, C.~Luo, X.~Geng, and
  B.~Guo, ``Magebench: Bridging large multimodal models to agents,''
  \emph{arXiv preprint arXiv:2412.04531}, 2024.

\bibitem{cot}
J.~Wei, X.~Wang, D.~Schuurmans, M.~Bosma, F.~Xia, E.~Chi, Q.~V. Le, D.~Zhou
  \emph{et~al.}, ``Chain-of-thought prompting elicits reasoning in large
  language models,'' \emph{Advances in neural information processing systems},
  vol.~35, pp. 24\,824--24\,837, 2022.

\bibitem{gpt3}
T.~Brown, B.~Mann, N.~Ryder, M.~Subbiah, J.~D. Kaplan, P.~Dhariwal,
  A.~Neelakantan \emph{et~al.}, ``Language models are few-shot learners,'' in
  \emph{Proc. Adv. Neural Inf. Process. Syst. (NeurIPS)}, vol.~33, 2020, pp.
  1877--1901.

\end{thebibliography}
\vfill

\end{document}